\documentclass[10pt,twocolumn,letterpaper]{article}

\usepackage[]{cvpr}

\usepackage{graphicx}
\usepackage{amsmath}
\usepackage{amssymb}
\usepackage{booktabs}
\usepackage{float}
\usepackage{color}
\usepackage{ulem}
\usepackage[pagebackref,breaklinks,colorlinks]{hyperref}

\usepackage[capitalize]{cleveref}
\crefname{section}{Sec.}{Secs.}
\Crefname{section}{Section}{Sections}
\Crefname{table}{Table}{Tables}
\crefname{table}{Tab.}{Tabs.}

\newcommand{\anonimize}[1]{\iftoggle{cvprfinal}{#1}{REDACTED}}

\begin{document}

\title{ \anonimize{BU}-CVKit: Extendable Computer Vision Framework for Species Independent Tracking and Analysis}
\author{Mahir Patel$^{1,2}$, Lucas Carstensen$^{2}$, Yiwen Gu$^{1}$, Michael Hasselmo$^2$, and Margrit Betke$^{1}$ \\ 
$^1$ Department of Computer Science and $^2$ Center for Systems Neuroscience \\
Boston University \\ }
\maketitle

\begin{abstract}
   A major bottleneck of interdisciplinary computer vision (CV) research is the lack of a framework that %
   eases the re-use and abstraction of state-of-the-art CV models  by CV and non-CV researchers alike.
   We present here \anonimize{BU}-CVKit, a computer vision framework that allows the creation of research pipelines with chainable Processors. The community can create plugins of their work for the framework, hence improving the re-usability, accessibility, and exposure of their work with minimal overhead. Furthermore, we provide \anonimize{MuSeqPose Kit}, a user interface for the pose estimation package of \anonimize{BU}-CVKit, which automatically scans for installed plugins and programmatically generates an interface for them based on the metadata provided by the user.
   It also provides software support for standard pose estimation features such as annotations, 3D reconstruction, re-projection, and camera calibration. Finally, we show examples of behavioral neuroscience pipelines created through the sample plugins created for our framework. %
\end{abstract}

\section{Introduction}
\label{sec:intro}
Computer Vision has the potential to become an integral part of interdisciplinary research. The advances in sub-fields such as Pose Estimation, Object Detection, Segmentation, and 3D Reconstruction can directly impact research in sciences as diverse as Conservation Ecology, Neuroscience, Physiology, Psychology, and many more. Integration of state-of-the-art computer vision with applied sciences can be supported by 
open-source python %
packages or wrappers like OpenCV ~\cite{opencv_library}, TensorFlow ~\cite{tensorflow}, PyTorch ~\cite{pytorch}, Scikit-learn ~\cite{sklearn}, OpenPose ~\cite{openpose}, MMPose ~\cite{mmpose2020}, DeepLabCut ~\cite{Nath476531}, and SLEAP ~\cite{Pereira2022} 
with different levels of abstraction and complexity.  However, researchers in the applied sciences usually have to rely on themselves to implement or, at best, adapt computer vision methods for their research pipeline. There is an absence of an extendable framework that provides a high-level abstraction for accessing the methods implemented through these open-source software. Such a framework could allow for easier integration of these methods in interdisciplinary pipelines, thus benefiting non-computer-vision specialists on the one hand and also providing computer vision researchers the tools to further their research on the other.

We present here \anonimize{BU}-CVKit framework, which bridges the accessibility gap to state-of-the-art computer vision research for researchers 
from diverse backgrounds and application disciplines.  The core idea behind \anonimize{BU}-CVKit  is to provide chainable modules that can be used to create abstract research pipelines which can be easily modified and shared with the community.
We also present \anonimize{MuSeqPose Kit}, which provides an intuitive user interface to the pose estimation sub-module of the framework.

We demonstrate the potential of our framework by implementing plugins for two state-of-the-art 2D/3D pose estimation methods, DeepLabCut~\cite{Nath476531} and OptiPose~\cite{patel_gu_carstensen_hasselmo_betke_2022}.
DeepLabCut is a widely-used feature-rich framework that provides state-of-the-art markerless 2D pose estimation of animals. OptiPose is a denoising auto-encoder that encodes postural dynamics to optimize coarse 3D poses. We use the plugins to create standard behavioral neuroscience pipelines. %

\section{\anonimize{BU}-CVKit Framework}
\anonimize{BU}-CVKit is an extendable framework that provides standard functionalities such as efficient input/output, evaluation metrics, geometric transformations, camera calibration, multi-view reconstruction, and other abstractions. 

To illustrate how \anonimize{BU}-CVKit can reduce programming overhead and thus potentially accelerate computer vision research, we give the following pipeline example. We denote a 2D pose estimation method as function~$f$, a 3D reconstruction method as function~$g$, a 3D pose filtering method as function~$h$, and a data analysis method as function~$i$.  \anonimize{BU}-CVKit enables the user to design a research pipeline that outputs
\begin{equation}
\label{eq:pipeline}
    o = i(h(g(f(data)))).
\end{equation}
Furthermore, consider a single-step 3D-pose estimation method $j$, where $j \approx g \ast f$. A user can replace $f$ and $g$ without affecting the semantics or flow of the pipeline and can thus explore different methods.

In the remainder of this section, we discuss the three major modules of the framework.

\begin{table}[b]
\caption{Average frames-per-second throughput of baseline OpenCV implementation and different \anonimize{BU}-CVKit buffered implementations, when computed over 1,000 1,024x1,024 H264-encoded frames under two CPU work modes on a 6-core Xeon processor.}
  \centering
{\small
\begin{tabular}{@{}lrr@{}}
    \toprule
    VideoReader & \multicolumn{2}{c}{Throughput [fps]} \\
    & CPU Load & CPU Idle \\
    \midrule 
    OpenCV VideoCapture(original) & 47.23&   162.07\\
    CVReader (buffered OpenCV) & 48.59  &  193.05 \\
    Deffcode & 49.30 &   193.09\\
    Decord & 64.71&   193.42\\
    \bottomrule
  \end{tabular}
  }
  \label{tab:perf_comp}
\end{table}

\subsection{Input/Output Modules}

Efficient and intuitive input/output is necessary for the application domains of computer vision.  We here describe two types of input/output modules. First, \anonimize{BU}-CVKit contains an abstract buffered VideoReader class that can be extended to provide sequential or random access to video frames using different backbone libraries. %
With our package, we provide
buffered implementation of OpenCV ~\cite{opencv_library}, Deffcode ~\cite{deffcode}, Decord, and an Image plugin,
which supports reading a directory of images and providing them as a video stream to support the data format adopted by the datasets. 
We compared the throughput of the standard 
 OpenCV video reader versus the buffered modules of \anonimize{BU}-CVKit under different CPU loads  reading high-resolution video frames, see Table~\ref{tab:perf_comp}.
The buffered module performs better than the standard OpenCV video readers and the improvement is much more evident when the CPU is idle.
This performance improvement may be beneficial for scientists who need to review videos of animal behavior and their analysis in real-time.
Users can also create their own IO plugins, perhaps one that utilizes GPU codecs to achieve even better performance.

The second I/O Module we describe here is the pose estimation data reader module.  It provides an abstract DataReader class that can be extended to provide intuitive access to 2D or 3D pose estimation data. Each instance of a person's or animal's pose is converted into a Skeleton object which in turn contains multiple Part Objects. Each Part object extends a Numpy array and therefore supports efficient vectorized operations. In addition, the Skeleton object supports further Pose-Estimation-specific features such as behavior annotations and unified arithmetic and geometric operations. 

To support input/output operations with a wide area of toolkits, the module includes implementations for reading, writing, and translating flattened CSV files, our custom $n$-dimensional CVKit files for storing pose data, HDF5~\cite{hdf5} files, Mat~\cite{MATLAB} files, and
2D DeepLabCut ~\cite{Nath476531} files.

\subsection{Processor Modules}

The pose estimation package provides an abstract Processor class that can be extended to implement plugins for state-of-the-art computer vision methods. The instances of these Processors are chainable and, therefore, can be used to create a pipeline that takes raw data and generate the desired output. The Processors can be classified into three categories - filters, generative, and utility. The filter package contains the Processors that denoise the input data. The included dimension-independent filters are a constant acceleration Kalman filter, linear interpolation, a statistical distance filter, a moving average filter, and a velocity filter.
The generative module contains the Processors that generate new data from the provided input. This module includes 3D reconstruction and reprojection processors, a kinematics generator, and plot generators. Finally, the utility module contains a file loader, a file saver, and an input statistics generator. 
Although these Processors are not directly used in processing or analyzing data, they are required to facilitate chaining and other utilities.

\subsection{Pose Estimation Utility Modules}

The framework provides several pose estimation utility modules. The camera calibration is a standard first step in any pose estimation work. Therefore we provide a calibration module that uses EasyWand~\cite{TheriaultFuJaBlEvWuBeHe14} to generate Direct Linear Transformation (DLT)~\cite{AbdelAziz-Karara-1971,kwon} coefficients and camera parameters. The 3D reconstruction and reprojection modules use DLT to estimate the 3D positions of matching 2D keypoints and mapping 3D coordinates into 2D camera planes. The metrics module for pose estimation provides the Mean Per Joint Position Error (MPJPE)~\cite{mono-3dhp2017} and dynamic percentage of correct keypoints (PCK@x)~\cite{mono-3dhp2017} metrics. Finally, the geometric transformations module enables 3D rotation, translation, and axes alignment.

\begin{figure}[t]
  \centering
  \includegraphics[width=0.9\linewidth]{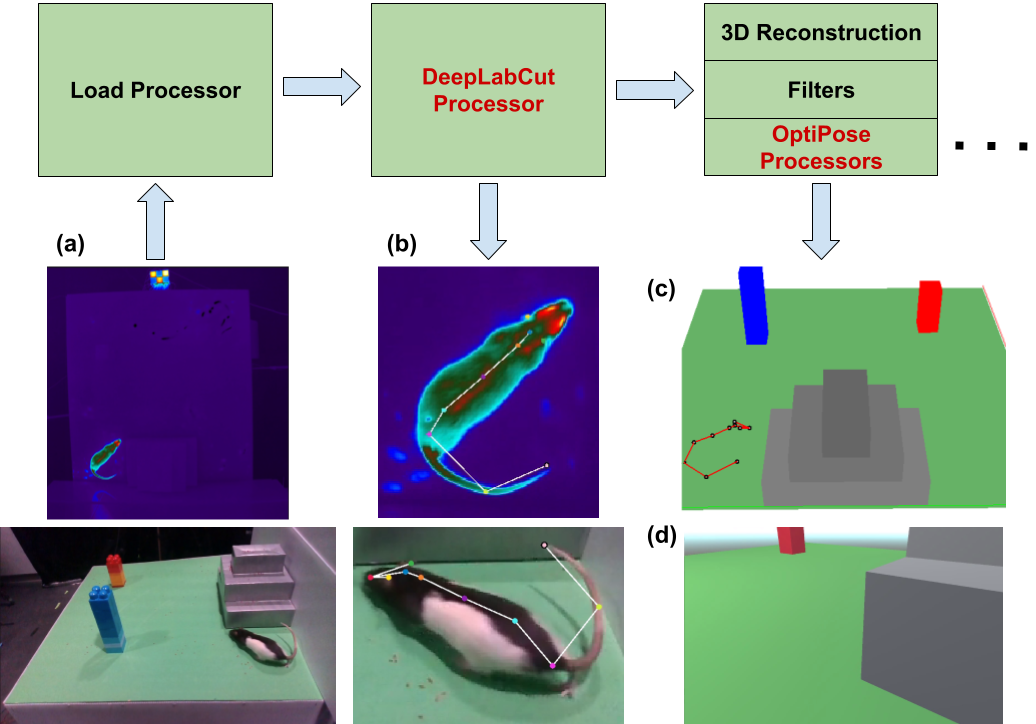}
  \caption{Chainable Processors: A Processor pipeline using the DeepLabCut ~\cite{Nath476531} and OptiPose ~\cite{patel_gu_carstensen_hasselmo_betke_2022} plugins. (a) The raw input thermal and RGB images. (b) Predicted 2D poses generated by the DeepLabCut plugin. (c) Optimized 3D poses generated by the pose-optimization Processor from the OptiPose plugin. (d) A simulated binocular egocentric view generated by the viewpoint reconstruction Processor from the OptiPose plugin.}
  \label{fig:plugin_chain}
\end{figure}

\section{Plugin Sub-system}
The abstract code design of the CVKit enables researchers  to extend the framework by creating plugins for their own work. The Processor class described in the previous section acts as the main entry point for plugins. We here showcase the potential of the plugin system by extending the capabilities of \anonimize{BU}-CVKit through our three plugins. Our DeepLabCut~\cite{Nath476531} plugin provides a Processor for running a DeepLabCut model on videos and providing 2D pose predictions. Our OptiPose~\cite{patel_gu_carstensen_hasselmo_betke_2022}  processor takes in the 3D poses generated by the reconstruction Processor and optimizes them through a pre-trained OptiPose model. It also supports generating a view direction vector that
can be fed into a game engine to generate a simulated egocentric view of the subject. 
The resulting pipeline created by \anonimize{BU}-CVKit is shown in
Fig.~\ref{fig:plugin_chain}. %
It starts with the DeepLabCut plugin to compute 2D predictions, the reconstruction  to perform 3D Reconstruction, a series of filters, pose optimization through an OptiPose model, and viewpoint simulation.

\begin{figure}[!t]
  \centering
  \includegraphics[height=0.6\linewidth]{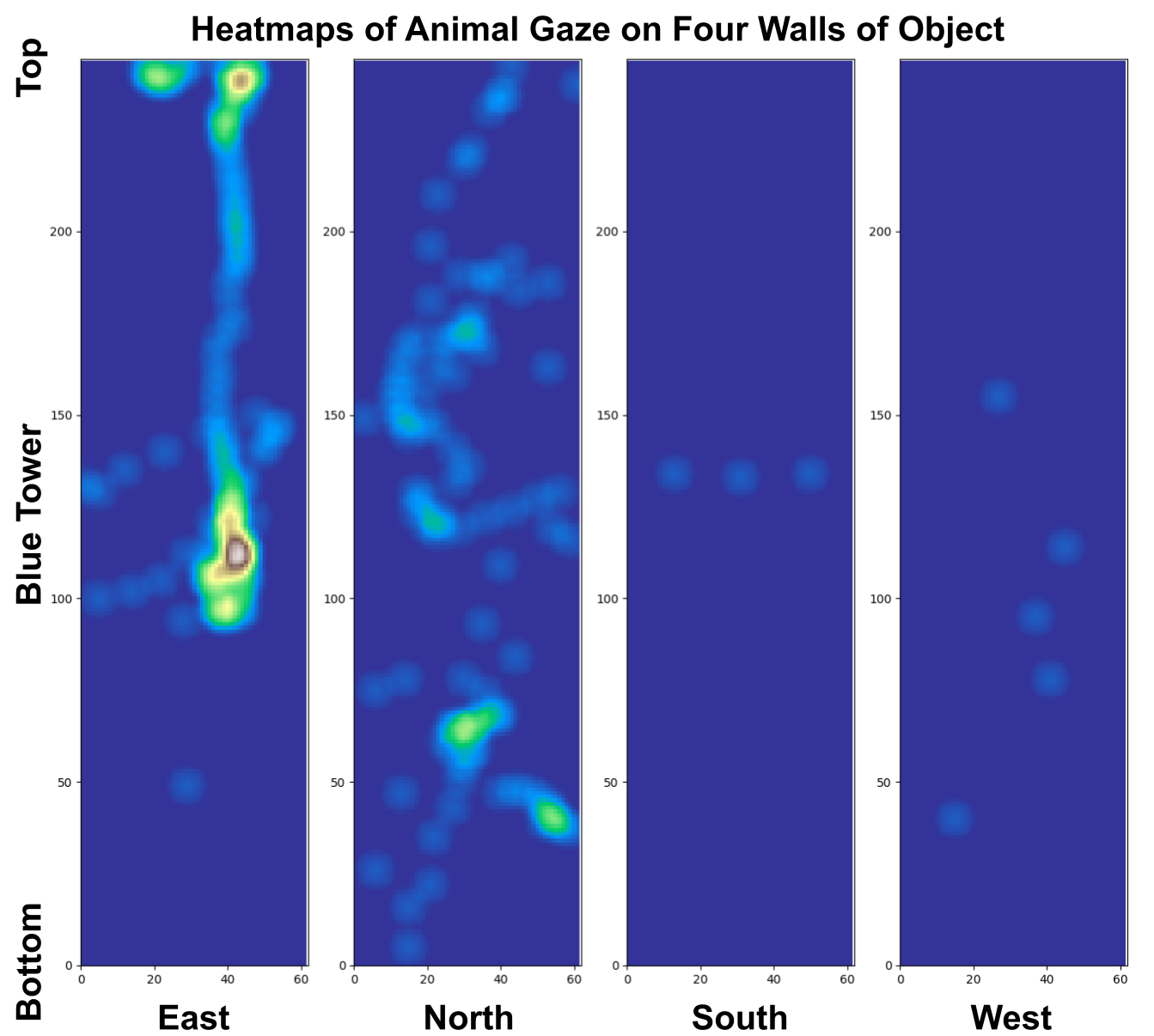}
  \caption{Gaze heatmaps computed by a Generative plugin that uses ray tracing with radially attenuated focus to approximate the gaze of the animal on the blue tower shown in Fig.~\ref{fig:plugin_chain}, for each wall of the tower (the west wall of the blue tower faces the red tower). The dimensions of the heatmaps correspond to the dimensions of the walls, i.e., 62~mm (width) and 247~mm (height). The rodent most often looked at the center of the East wall (hot colors).  
  } 
  \label{fig:obj_gaze}
  \vspace{-0.3cm} 
\end{figure}
\begin{figure}[!t]
  \centering
  \includegraphics[width=0.9\linewidth]{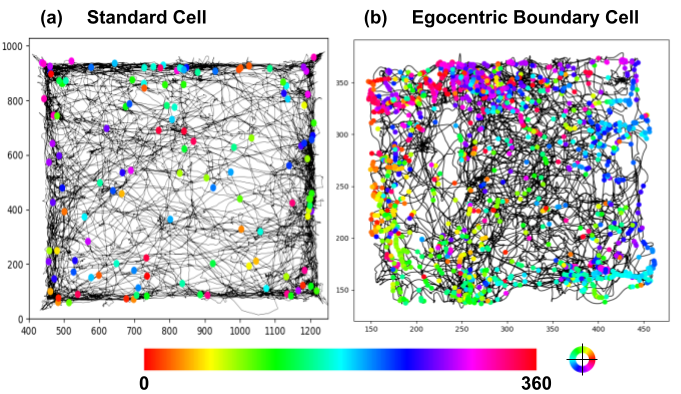}
  \caption{Results of a Generative plugin that combines neural recording with animal tracking to plot cellular activity and their corresponding spatial locations. The black lines indicate the trajectory of the rodent exploring an area (as in Fig.\ref{fig:plugin_chain} but without objects),
  and each dot is the location where the cell fired. The color of the dot represents the head direction of the rodent at the time the cell fired. (a) Example of non-deterministic cell firing. (b) 
  Example of an Egocentric Boundary Cell firing. }
  \label{fig:cell}
  \vspace{-0.3cm} 
\end{figure}

\begin{figure*}[!ht]
  \centering
  \includegraphics[width=0.80\linewidth]{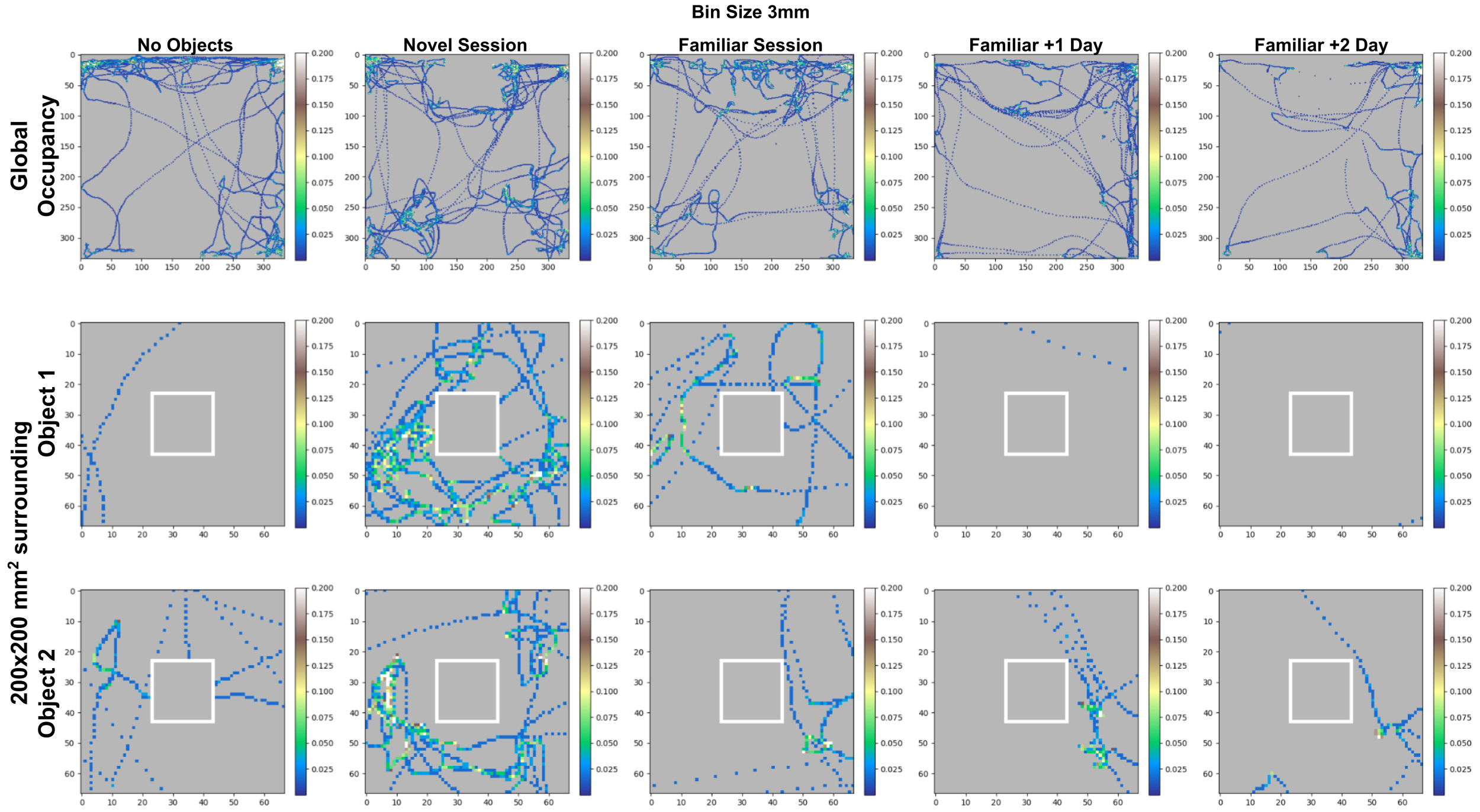}
  \includegraphics[width=0.80\linewidth]{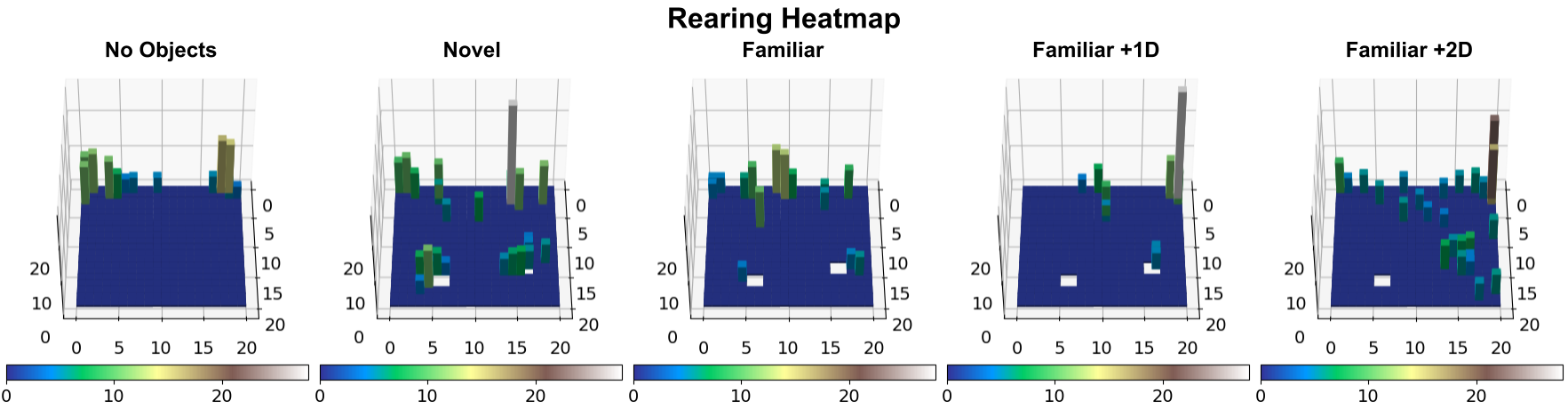}
  \caption{Top: Results of a behavioral analysis plugin that generates Occupancy Maps for comparing the change in behavior corresponding to the novelty of objects. Bottom: Results of a behavioral analysis plugin that generates "rearing" plots to compare exploratory behavior under novel or familiar conditions. } %
  \label{fig:occupancy}
  \vspace{-0.3cm} 
\end{figure*}

The NeuroAnalysis plugin  processes the tracking data generated from the OptiPose plugin (not shown in Fig.~\ref{fig:plugin_chain}). It provides a single-ray-tracing Processor that is used by other behavioral analysis generators. 
With the plugin, the gaze direction of the animal can be analyzed by tracing the view direction vector to surfaces in the animal's environment. Here we make the assumption that the rodent is generally looking at the direction it is facing, i.e., where its snout is pointing.  We also model the rodent's attention focus, where the attention score attenuates radially outwards from the assumed gaze point.  With these tools, we can localize the rodent's approximate focus of interest on the four walls of the blue tower, shown in Fig.~\ref{fig:plugin_chain}. 
Summative analysis then leads to heat maps that highlight to which areas on the tower the animal seemed to have paid attention most often, see Fig.~\ref{fig:obj_gaze}.

The EBC-Processor helps detect Egocentric Boundary Cells (EBC), %
neurons that exhibit increased firing when a boundary is at a specific distance and direction relative to the subject ~\cite{alexander2020egocentric}. Analytical plots, as in Fig.~\ref{fig:cell}, are generally used to distinguish non-EBC patterns from EBCs. Each plot is produced for a specific cell. The plot on the left in Fig.~\ref{fig:cell} shows no discernible directional or spatial pattern in the cell activity, whereas the plot on the right exemplifies a firing pattern
when the boundary is on the left of the rodent. 

Animal behavioral analysis studies how an animal explores an environment under a spectrum of novel conditions. The amount of time the rodent spends occupying certain regions and the instances of rearing are used as indicators of exploration and memory~\cite{lcdissertation,lueptow}. The Occupancy Processor can generate 2D plots that show the distribution of where the time was spent globally as well as locally near the objects of interest (Fig.~\ref{fig:occupancy}). Similarly, the Rearing Processor detects rearing instances by monitoring selected keypoints to generate 3D bar plots of rearing frequency (Fig.~\ref{fig:occupancy}).

We stress that a major advantage of the framework architecture is that the processors are easily replaceable. The DeepLabCut Processor and the OptiPose Processor can be replaced by any 2D pose estimator and 3D pose optimizer, respectively, without affecting the neuro-processors. They can also be bypassed entirely by creating a plugin for one of the single-stage 3D pose estimators. This allows researchers to try different methods without drastically changing their analysis pipeline.

\section{\anonimize{MuSeqPose Kit}} %

We provide \anonimize{MuseqPose Kit}, a user interface to the pose-estimation sub-package of \anonimize{BU}-CVKit. It provides a powerful video annotation widget that supports interpolating annotations across frames. The widget also provides a reprojection toolbox to mitigate the need for annotating every view. The camera calibration widget automatically picks a diverse set of synchronized annotated frames and generates the necessary files for the EasyWand Calibration package. The synchronized playback widget allows for displaying videos with real-time plots. Finally, the Pipeline widget automatically scans for the installed plugin metadata and generates a user interface to access the underlying Processors. Using this widget, users can create, visualize, execute, and save their research pipelines using this widget.

\section{Conclusions}

With \anonimize{BU}-CVKit Framework, we have provided an extendable framework for simplifying and accelerating interdisciplinary research. We showcase the usability of such a framework by implementing plugins for state-of-the-art computer vision research  and using them for behavioral neuroscience. 
We plan to add extendable Segmentation and Object Detection packages.
The plug-and-play aspect of the Processors enables researchers to try different methods without extensively modifying their pipeline. Finally, since non-computer-science researchers may have a varying range of programming skills, we provide a powerful user interface that automatically adapts to the installed plugins while providing pose estimation features.

\newpage
{\small
\bibliographystyle{ieee_fullname}
\bibliography{egbib,animal-tracking,Betke-publications}

\begin{thebibliography}{10}\itemsep=-1pt

\bibitem{tensorflow}
Mart\'{i}n Abadi, Ashish Agarwal, Paul Barham, Eugene Brevdo, Zhifeng Chen,
  Craig Citro, Greg~S. Corrado, Andy Davis, Jeffrey Dean, Matthieu Devin,
  Sanjay Ghemawat, Ian Goodfellow, Andrew Harp, Geoffrey Irving, Michael Isard,
  Yangqing Jia, Rafal Jozefowicz, Lukasz Kaiser, Manjunath Kudlur, Josh
  Levenberg, Dandelion Man\'{e}, Rajat Monga, Sherry Moore, Derek Murray, Chris
  Olah, Mike Schuster, Jonathon Shlens, Benoit Steiner, Ilya Sutskever, Kunal
  Talwar, Paul Tucker, Vincent Vanhoucke, Vijay Vasudevan, Fernanda Vi\'{e}gas,
  Oriol Vinyals, Pete Warden, Martin Wattenberg, Martin Wicke, Yuan Yu, and
  Xiaoqiang Zheng.
\newblock {TensorFlow}: Large-scale machine learning on heterogeneous systems,
  2015.
\newblock Software available from tensorflow.org.

\bibitem{AbdelAziz-Karara-1971}
Y.~I. Abdel-Aziz and H.~M. Karara.
\newblock Direct linear transformation from comparator coordinates into object
  space coordinates in close-range photogrammetry.
\newblock {\em Proceedings of the Symposium on Close-Range Photogrammetry},
  pages 1--18, 1971.

\bibitem{alexander2020egocentric}
Andrew~S. Alexander, Lucas~C. Carstensen, James~R. Hinman, Florian Raudies,
  G.~William Chapman, and Michael~E. Hasselmo.
\newblock Egocentric boundary vector tuning of the retrosplenial cortex.
\newblock {\em Science advances}, 6(8):eaaz2322, 2020.

\bibitem{opencv_library}
G. Bradski.
\newblock {The OpenCV Library}.
\newblock {\em Dr. Dobb's Journal of Software Tools}, 2000.

\bibitem{openpose}
Z. {Cao}, G. {Hidalgo Martinez}, T. {Simon}, S. {Wei}, and Y.~A. {Sheikh}.
\newblock Openpose: Realtime multi-person 2d pose estimation using part
  affinity fields.
\newblock {\em IEEE Transactions on Pattern Analysis and Machine Intelligence},
  2019.

\bibitem{lcdissertation}
Lucas~C. Carstensen.
\newblock {\em Neural representations of environmental features in
  retrosplenial cortex and 3-dimensional reconstruction of animal pose}.
\newblock PhD thesis, 2023.

\bibitem{mmpose2020}
MMPose Contributors.
\newblock Openmmlab pose estimation toolbox and benchmark.
\newblock \url{https://github.com/open-mmlab/mmpose}, 2020.

\bibitem{hdf5}
The~{HDF} {G}roup.
\newblock Hierarchical data format version 5, 2000--2010.

\bibitem{lueptow}
Lindsey~M. Lueptow.
\newblock Novel object recognition test for the investigation of learning and
  memory in mice.
\newblock {\em Journal of Visualized Experiments: JoVE}, 55718:126, 2017.

\bibitem{MATLAB}
The {M}ath{W}orks {I}nc.
\newblock Matlab, 2023.

\bibitem{mono-3dhp2017}
Dushyant Mehta, Helge Rhodin, Dan Casas, Pascal Fua, Oleksandr Sotnychenko,
  Weipeng Xu, and Christian Theobalt.
\newblock Monocular 3d human pose estimation in the wild using improved cnn
  supervision.
\newblock In {\em 3D Vision (3DV), 2017 Fifth International Conference on}.
  IEEE, 2017.

\bibitem{Nath476531}
Tanmay Nath, Alexander Mathis, An~Chi Chen, Amir Patel, Matthias Bethge, and
  Mackenzie~Weygandt Mathis.
\newblock Using {D}eep{L}ab{C}ut for 3{D} markerless pose estimation across
  species and behaviors.
\newblock {\em bioRxiv}, 2018.

\bibitem{pytorch}
Adam Paszke, Sam Gross, Francisco Massa, Adam Lerer, James Bradbury, Gregory
  Chanan, Trevor Killeen, Zeming Lin, Natalia Gimelshein, Luca Antiga, Alban
  Desmaison, Andreas Köpf, Edward Yang, Zach DeVito, Martin Raison, Alykhan
  Tejani, Sasank Chilamkurthy, Benoit Steiner, Lu Fang, Junjie Bai, and Soumith
  Chintala.
\newblock Pytorch: An imperative style, high-performance deep learning library,
  2019.

\bibitem{patel_gu_carstensen_hasselmo_betke_2022}
Mahir Patel, Yiwen Gu, Lucas~C. Carstensen, Michael~E. Hasselmo, and Margrit
  Betke.
\newblock Animal pose tracking: 3d multimodal dataset and token-based pose
  optimization.
\newblock {\em International Journal of Computer Vision}, 131(2):514–530,
  2022.

\bibitem{sklearn}
F. Pedregosa, G. Varoquaux, A. Gramfort, V. Michel, B. Thirion, O. Grisel, M.
  Blondel, P. Prettenhofer, R. Weiss, V. Dubourg, J. Vanderplas, A. Passos, D.
  Cournapeau, M. Brucher, M. Perrot, and E. Duchesnay.
\newblock Scikit-learn: Machine learning in {P}ython.
\newblock {\em Journal of Machine Learning Research}, 12:2825--2830, 2011.

\bibitem{Pereira2022}
Talmo~D. Pereira, Nathaniel Tabris, Arie Matsliah, David~M. Turner, Junyu Li,
  Shruthi Ravindranath, Eleni~S. Papadoyannis, Edna Normand, David~S. Deutsch,
  Z.~Yan Wang, Grace~C. McKenzie-Smith, Catalin~C. Mitelut, Marielisa~Diez
  Castro, John D'Uva, Mikhail Kislin, Dan~H. Sanes, Sarah~D. Kocher, Samuel
  S.-H. Wang, Annegret~L. Falkner, Joshua~W. Shaevitz, and Mala Murthy.
\newblock {SLEAP}: A deep learning system for multi-animal pose tracking.
\newblock {\em Nature Methods}, Apr. 2022.

\bibitem{deffcode}
Abhishek~Singh Thakur.
\newblock abhitronix/deffcode: v0.2.4, Oct. 2022.

\bibitem{TheriaultFuJaBlEvWuBeHe14}
D.~H. Theriault, N.~W. Fuller, B.~E. Jackson, E. Bluhm, D. Evangelista, Z. Wu,
  M. Betke, and T.~L. Hedrick.
\newblock A protocol and calibration method for accurate multi-camera field
  videography.
\newblock {\em The Journal of Experimental Biology}, 217:1843--1848, Feb. 2014.
\newblock Open access online,
  \href{http://jeb.biologists.org/content/early/2014/02/20/jeb.100529.abstract.html?papetoc">}{pdf.}

\bibitem{kwon}
Kwon Young-Hoo.
\newblock Dlt method.
\newblock \url{http://www.kwon3d.com/theory/dlt/dlt.html}.
\newblock Accessed April 28, 2023.

\end{thebibliography}
}

\end{document}